\newif\ifJOURNAL
\JOURNALfalse
\newif\ifarXiv
\arXivfalse
\newif\ifWP
\WPfalse
\newif\ifFULL
\FULLfalse

\arXivtrue

\newif\ifTR		
\TRfalse
\ifarXiv\TRtrue\fi
\ifWP\TRtrue\fi

\newif\ifnotJOURNAL	
\notJOURNALtrue
\ifJOURNAL\notJOURNALfalse\fi

\ifJOURNAL
\documentclass[twoside,11pt]{article}

\usepackage{jmlr2e,amsmath,amsfonts,latexsym}

\jmlrheading{0}{2000}{00-00}{9/05}{00/00}{Vladimir Vovk}

\ShortHeadings{Competitive on-line learning}{Vovk}
\firstpageno{1}
\newcommand{\Extra}[1]{}
\fi

\ifarXiv
\documentclass{article}
\usepackage{amsmath,amsfonts,amssymb,latexsym}
\usepackage[dvips]{graphicx}
\newcommand{\Extra}[1]{}
\fi

\ifWP
\documentclass[toc]{gtarticle}
\usepackage{amsmath,amsfonts,latexsym,epsfig}
\renewcommand{\Extra}[1]{}
\fi

\ifFULL
\documentclass{article}
\usepackage{amsmath,amsfonts,latexsym}
\newcommand{\Extra}[1]{\begingroup\sl #1\endgroup}
\fi

\allowdisplaybreaks[3]

\ifJOURNAL
  \newcommand{\GTPVIII}{vovk/etal:2005AIStats}
  \newcommand{\GTPX}{vovk/etal:2005ALT-GTP10}
  \newcommand{\GTPXIII}{vovk:2005ALT-GTP13}
  \newcommand{\GTPXIV}{vovk:2005ALT-GTP14}
\fi
\ifarXiv
  \newcommand{\GTPVIII}{GTP8arXiv}
  \newcommand{\GTPX}{GTP10arXiv}
  \newcommand{\GTPXIII}{GTP13arXiv}
  \newcommand{\GTPXIV}{GTP14arXiv-local}
\fi
\ifWP
  \newcommand{\GTPVIII}{GTP8}
  \newcommand{\GTPX}{GTP10}
  \newcommand{\GTPXIII}{GTP13}
  \newcommand{\GTPXIV}{GTP14}
\fi
\ifFULL
  \newcommand{\GTPVIII}{GTP8arXiv}
  \newcommand{\GTPX}{GTP10arXiv}
  \newcommand{\GTPXIII}{GTP13arXiv}
  \newcommand{\GTPXIV}{GTP14arXiv}
\fi

\newcommand{\Vladimir}{Vladimir }
\newcommand{\DOT}{.}

\newcommand{\bbbr}{\mathbb{R}}		

\newcommand{\st}{\mathop{|}}

\newcommand{\D}{\,\mathrm{d}}

\newcommand{\K}{\mathcal{K}}		
\newcommand{\KKK}{\mathbf{K}}		
\newcommand{\FFF}{\mathcal{F}}		
\newcommand{\HHH}{\mathcal{H}}		
\newcommand{\SSS}{\mathcal{S}}		
\newcommand{\CC}{\mathbf{c}}		

\newcommand{\bbbe}{\mathbb{E}}		
\newcommand{\Expect}{\mathop{\bbbe}\nolimits}
\newcommand{\Exp}{\mathop{\mathrm{Exp}}\nolimits}

\newcommand{\sign}{\mathop{\mathrm{sign}}\nolimits}
\newcommand{\dom}{\mathop{\mathrm{dom}}\nolimits}

\newcommand{\poundsstar}{\pounds\mspace{-1.1mu}^{\dagger}}
\newcommand{\starpounds}{{}^{\dagger}\!\!\pounds}
\newcommand{\starpoundsstar}{{}^{\dagger}\!\!\pounds\mspace{-1.1mu}^{\dagger}}

\ifnotJOURNAL
\newtheorem{theorem}{Theorem}
\newtheorem{lemma}{Lemma}
\newtheorem{corollary}{Corollary}

\newtheorem{remark}{Remark}
\newenvironment{proof}
  {\trivlist\item[\hskip\labelsep\textbf{Proof}]}
  {\endtrivlist}
\newcommand{\BlackBox}{\rule{.3em}{1.5ex}}
\fi
\renewenvironment{proof}
  {\trivlist\item[\hskip\labelsep\textbf{Proof}]}
  {\endtrivlist}
\newenvironment{Proof}[1]
  {\trivlist\item[\hskip\labelsep\textbf{Proof #1}]}
  {\endtrivlist}
\newcommand{\qedtext}{\unskip\nobreak\hfil
  \penalty50\hskip1em\null\nobreak\hfil\BlackBox
  \parfillskip=0pt\finalhyphendemerits=0\endgraf}
\newcommand{\qedmath}{\tag*{\BlackBox}}
\newenvironment{Remark}
  {\begin{remark}\begingroup\rm\relax}
  {\endgroup\end{remark}}
\newenvironment{remark*}
  {\trivlist\item[\hskip\labelsep{\bfseries Remark}]\relax}
  {\endtrivlist}

\ifnotJOURNAL
  \newcommand{\citealt}[1]{\cite{#1}}
  \newcommand{\citet}[1]{\cite{#1}}
  \newcommand{\citep}[1]{\cite{#1}}
\fi

\ifJOURNAL
  \newcommand{\Uppercase}[1]{\uppercase{#1}}
\fi
\ifnotJOURNAL
  \newcommand{\Uppercase}[1]{#1}
\fi

\newlength{\IndentI}
\newlength{\IndentII}
\newlength{\IndentIII}
\newlength{\WidthI}
\newlength{\WidthII}
\newlength{\WidthIII}
\ifJOURNAL
\title{Competitive on-line learning
  with a convex loss function}

\author{\name Vladimir Vovk \email vovk@cs.rhul.ac.uk\\
  \addr Computer Learning Research Centre, Department of Computer Science\\
  Royal Holloway, University of London, Egham, Surrey TW20 0EX, UK}

\editor{000}
\fi

\ifarXiv
\title{Competitive on-line learning\\
  with a convex loss function}

\author{Vladimir Vovk\\
\texttt{vovk\textrm{@}cs.rhul.ac.uk}\\
\texttt{http://vovk.net}}
\fi

\ifWP
\title{Competitive on-line learning
  with a convex loss function}

\author{Vladimir Vovk}

\twodatestrue

\fi

\ifFULL
\title{Competitive on-line learning
  with a convex loss function}
\author{Vladimir Vovk}
\fi

\begin{document}
\maketitle

\begin{abstract}
  We consider the problem of sequential decision making under uncertainty
  in which the loss caused by a decision depends on the following binary observation.
  In competitive on-line learning,
  the goal is to design decision algorithms
  that are almost as good as the best decision rules
  in a wide benchmark class,
  without making any assumptions about the way the observations are generated.
  However, standard algorithms in this area can only deal with finite-dimensional
  (often countable)
  benchmark classes.
  In this paper we give similar results
  for decision rules
  ranging over an arbitrary reproducing kernel Hilbert space.
  For example, it is shown that
  for a wide class of loss functions
  (including the standard square, absolute, and log loss functions)
  the average loss of the master algorithm, over the first $N$ observations,
  does not exceed the average loss of the best decision rule
  with a bounded norm plus $O(N^{-1/2})$.
  Our proof technique is very different from the standard ones
  and is based on recent results about defensive forecasting.
  Given the probabilities produced by a defensive forecasting algorithm,
  which are known to be well calibrated and to have good resolution in the long run,
  we use the expected loss minimization principle
  to find a suitable decision.
\end{abstract}

\ifJOURNAL
\fi

\section{Introduction}

In the simple problem of sequential decision making
that we consider in this paper,
the loss $\lambda(y_n,\gamma_n)$
(maybe negative)
caused by a decision $\gamma_n$
depends only on the following binary observation $y_n$.
All relevant information available to the decision maker
by the time he makes his decision
is collected in what we call the datum, $x_n$.
For example, in time series applications the datum may contain
all or the most recent observations;
in pattern recognition,
where the observations are the true classes of patterns,
the datum may be the vector of a pattern's attributes.

The traditional approach to this problem
assumes a statistical model for the sequence
of pairs $(x_n,y_n)$;
e.g., statistical learning theory (\citealt{vapnik:1998})
assumes that the $(x_n,y_n)$ are generated independently
from the same probability distribution.
A more recent approach,
known in learning theory as ``prediction with expert advice''
(e.g., \citealt{cesabianchi/etal:1997})
and in information theory as ``universal prediction''
(e.g., \citealt{feder/etal:1992,merhav/feder:1998}),
avoids making assumptions about the way
the observations and data are generated.
Instead,
the goal of the decision maker is to compete
with a more or less general benchmark class of decision rules,
mapping the $x$s to the $y$s
(the framework of prediction with expert advice is usually even more general).
We will use the phrase ``competitive on-line'' to refer to this area
(as in \citealt{vovk:2001competitive},
emphasizing similarities to competitive on-line algorithms
in computation theory).

First papers on competitive on-line learning
with general loss functions
(e.g., \citealt{cesabianchi/etal:1997,vovk:1990})
dealt with countable (often finite) benchmark classes.
The next step was to consider finite-dimensional benchmark classes
(e.g., \citealt{freund:1996,kivinen/warmuth:1997,vovk:2001competitive}).
This paper continues with infinite-dimensional classes.
(Such classes were considered earlier
\ifJOURNAL by \citealt{kimber/long:1995} and \citealt{long:2000}\fi
\ifnotJOURNAL by Kimber and Long \cite{kimber/long:1995,long:2000}\fi,
who, however, assumed that the benchmark class
contains a perfect decision rule.)
To get an idea of our central results,
the reader is advised to start from Corollaries
\ifJOURNAL\ref{cor:square}, \ref{cor:absolute}, and \ref{cor:log}.\fi
\ifnotJOURNAL\ref{cor:square}--\ref{cor:log}.\fi

Our implicit assumption,
common with other work in competitive on-line learning,
is that the decision maker is ``small'':
his decisions do not affect the future observations.
This is not a mathematical assumption:
as already mentioned,
we do not make any assumptions at all about the way observations are generated;
however,
interpretation of our results becomes problematic
if the decision maker is not small.
``Big'' decision makers can still use our algorithms
for prediction
(cf.\ Remarks \ref{rem:prediction} and \ref{rem:illustration} below).

In conclusion of this section
we will briefly describe the content of the paper.
Our main result is stated in \S\ref{sec:main},
and several examples are given in \S\ref{sec:examples}.
It is proved in \S\ref{sec:proof};
in \S\ref{sec:idea} we describe the main ideas behind the proof
and in \S\ref{sec:ALN}
we prove some preparatory results for \S\ref{sec:proof}.
Our decision algorithm is explicitly described in \S\ref{sec:algorithm}.
We conclude with a short list of directions of further research
(\S\ref{sec:discussion}).
A preliminary version of this paper is to appear as \citet{\GTPXIV}.
In this new version we made the title of the paper more specific
(the old title was even somewhat misleading:
in prediction with expert advice,
the experts are usually completely free
in making their decisions).

\section{Main \Uppercase{r}esult}
\label{sec:main}

Our decision protocol is:

\bigskip

\parshape=5
\IndentI  \WidthI
\IndentII \WidthII
\IndentII \WidthII
\IndentII \WidthII
\IndentI  \WidthI
\noindent
FOR $n=1,2,\dots$:\\
  Reality announces $x_n\in\mathbf{X}$.\\
  Decision Maker announces $\gamma_n\in\Gamma$.\\
  Reality announces $y_n\in\{0,1\}$.\\
END FOR.

\bigskip

\noindent
At each step (or \emph{round}) $n$
Decision Maker makes a decision $\gamma_n$
whose consequences depend on the \emph{observation} $y_n\in\{0,1\}$
chosen by Reality.
All relevant information available to Decision Maker by the time he makes his decision
is collected in $x_n$,
called the \emph{datum}.
We assume that the data $x_n$ are elements
of a \emph{data space} $\mathbf{X}$
and that the decisions are elements of a \emph{decision space} $\Gamma$
(both sets assumed non-empty).

\begin{Remark}\label{rem:prediction}
  In this paper we are interested, first of all, in prediction of future observations.
  However,
  our framework allows a fairly wide class of loss functions,
  not all of which can be interpreted in terms of predictions
  (such as, e.g., Cover's and long-short games,
  in the terminology of \citealt{vovk:2001competitive}, \S2).
  This is the main reason why we prefer to talk about decision making in general;
  another reason is that in \S\ref{sec:ALN}
  we will deal with a very different kind of prediction
  (for which we reserve the term ``forecasting'').
\end{Remark}

A \emph{decision strategy} is a strategy for Decision Maker
in this protocol
(explicitly defined specific strategies will also be called
``decision algorithms'').
Its performance is measured with a \emph{loss function}
$\lambda:\{0,1\}\times\Gamma\to\bbbr$,
and so its cumulative loss over the first $N$ rounds is
\begin{equation*}
  \sum_{n=1}^N
  \lambda(y_n,\gamma_n).
\end{equation*}
The pair $(\Gamma,\lambda)$ is the \emph{game} being played.
Decision Maker will compete against a class $\FFF$,
called the \emph{benchmark class},
of functions $D:\mathbf{X}\to\Gamma$
considered as decision rules;
the cumulative loss suffered by such a decision rule is
\begin{equation*}
  \sum_{n=1}^N
  \lambda(y_n,D(x_n)).
\end{equation*}

Before stating our main result we define some useful notions
connected with the two main components of our decision framework,
the game $(\Gamma,\lambda)$ and the benchmark class $\FFF$.
The reader might want in parallel to read the next section,
which describes some important examples of games and benchmark classes.

\subsection*{Games}

The \emph{exposure} $\Exp_{\lambda}(\gamma)\in\bbbr$ of a decision $\gamma\in\Gamma$ is
\begin{equation*}
  \Exp_{\lambda}(\gamma)
  :=
  \lambda(1,\gamma) - \lambda(0,\gamma)
\end{equation*}
and the \emph{exposure} $\Exp_{\lambda,D}:\mathbf{X}\to\bbbr$ of a decision rule $D$
at a point $x\in\mathbf{X}$ is
\begin{equation*}
  \Exp_{\lambda,D}(x)
  :=
  \lambda(1,D(x)) - \lambda(0,D(x)).
\end{equation*}
Let $\lambda(p,\gamma)$ be the expected loss
caused by taking a decision $\gamma$ when the probability of $1$ is $p$:
\begin{equation}\label{eq:expected-loss}
  \lambda(p,\gamma)
  :=
  p \lambda(1,\gamma)
  +
  (1-p) \lambda(0,\gamma).
\end{equation}

We only consider games $(\Gamma,\lambda)$ such that
\begin{equation}\label{eq:C-01}
  C_0
  :=
  \inf_{\gamma\in\Gamma}
  \lambda(0,\gamma),
  \quad
  C_1
  :=
  \inf_{\gamma\in\Gamma}
  \lambda(1,\gamma)
\end{equation}
are finite.
It is convenient
(see, e.g., \citealt{kalnishkan/vyugin:2005})
to summarize a game by its \emph{superdecision set}
\begin{equation}\label{eq:super}
  \Sigma
  :=
  \left\{
    (x,y)\in\bbbr^2
    \st
    \exists\gamma\in\Gamma:
    x\ge\lambda(0,\gamma)
    \text{ and }
    y\ge\lambda(1,\gamma)
  \right\};
\end{equation}
elements of this set will be called \emph{superdecisions}.
Superdecisions of the form 
$(\lambda(0,\gamma),\lambda(1,\gamma))$
will sometimes be called \emph{decisions}.
We will assume, additionally,
that the set $\Sigma\subseteq\bbbr^2$ is convex and closed.
The \emph{Eastern tail} of the game is the function
\begin{equation}\label{eq:Eastern}
\begin{split}
  f:[C_0,\infty)&\to\bbbr\cup\{\infty\}\\
    x&\mapsto\inf\{y\st(x,y)\in\Sigma\} - C_1
\end{split}
\end{equation}
and its \emph{Northern tail} is
\begin{equation}\label{eq:Northern}
\begin{split}
  g:[C_1,\infty)&\to\bbbr\cup\{\infty\}\\
    y&\mapsto\inf\{x\st(x,y)\in\Sigma\} - C_0,
\end{split}
\end{equation}
where, as usual, $\inf\emptyset:=\infty$;
it is clear that $f$ and $g$ are nonnegative everywhere
and finite on $(C_0,\infty)$ and $(C_1,\infty)$,
respectively.

\subsection*{The \Uppercase{t}heorem}

A \emph{reproducing kernel Hilbert space} (RKHS) on $\mathbf{X}$
is a Hilbert space $\FFF$ of real-valued functions on $\mathbf{X}$
such that the evaluation functional $f\in\FFF\mapsto f(x)$
is continuous for each $x\in\mathbf{X}$.
By the Riesz--Fischer theorem,
for each $x\in\mathbf{X}$ there exists a function $\KKK_x\in\FFF$ such that
\begin{equation*}
  f(x)
  =
  \langle \KKK_x,f\rangle_{\FFF},
  \quad
  \forall f\in\FFF.
\end{equation*}
Let
\begin{equation}\label{eq:C}
  \CC_{\FFF}
  :=
  \sup_{x\in\mathbf{X}}
  \left\|\KKK_x\right\|_{\FFF};
\end{equation}
we will be interested in the case $\CC_{\FFF}<\infty$.
With each game $(\Gamma,\lambda)$ and each RKHS $\FFF$
we associate the non-negative (but maybe infinite) constant
$\CC_{\lambda,\FFF}$ defined by
\begin{equation}\label{eq:big-C}
\begin{aligned}
  \CC^2_{\lambda,\FFF}
  &:=
  \sup_{p\in(0,1)}
  \sup_{\gamma\in\Gamma_p}
  \sup_{x\in\mathbf{X}}
  p(1-p)
  \left(
    \Exp_{\lambda}^2(\gamma)
    +
    \|\KKK_x\|_{\FFF}^2
  \right)\\
  &=
  \sup_{p\in(0,1)}
  \sup_{\gamma\in\Gamma_p}
  p(1-p)
  \left(
    \Exp_{\lambda}^2(\gamma)
    +
    \CC_{\FFF}^2
  \right),
\end{aligned}
\end{equation}
where
$
  \Gamma_p
  :=
  \arg\min_{\gamma\in\Gamma}
  \lambda(p,\gamma)
$
(and $\lambda(p,\gamma)$ is defined by (\ref{eq:expected-loss})).

The following is our main result.
\begin{theorem}\label{thm:main}
  Let the game $(\Gamma,\lambda)$ be such that
  (\ref{eq:C-01}) are finite,
  the superdecision set $\Sigma$ is convex and closed,
  and the tails $f$ and $g$ satisfy
  \begin{equation}\label{eq:tails1}
    f'_{+}(t)=O(t^{-2}),
    \quad
    g'_{+}(t)=O(t^{-2})
  \end{equation}
  as $t\to\infty$,
  where $f'_{+}$ and $g'_{+}$ stand for the right derivatives
  (see, e.g., \citealt{rockafellar:1970}, \S23)
  of $f$ and $g$.
  Let $\FFF$ be an RKHS on $\mathbf{X}$
  and $\CC_{\FFF}$, $\CC_{\lambda,\FFF}$ be defined by (\ref{eq:C}) and (\ref{eq:big-C}).
  Suppose $\CC_{\FFF}<\infty$.
  Then $\CC_{\lambda,\FFF}<\infty$
  and there is a decision strategy which guarantees that
  \begin{equation}\label{eq:goal}
    \sum_{n=1}^N
    \lambda(y_n,\gamma_n)
    \le
    \sum_{n=1}^N
    \lambda(y_n,D(x_n))
    +
    \CC_{\lambda,\FFF}
    \left(
      \left\|\Exp_{\lambda,D}\right\|_{\FFF}+1
    \right)
    \sqrt{N}
  \end{equation}
  for all $N=1,2,\ldots$ and all $D:\mathbf{X}\to\Gamma$ with $\Exp_{\lambda,D}\in\FFF$.
\end{theorem}

\begin{Remark}
  If the loss function $\lambda$ is bounded,
  (\ref{eq:tails1}) holds trivially.
  The right derivatives in (\ref{eq:tails1})
  can be replaced by the corresponding left derivatives,
  since $\left|f'_{+}\right|\le\left|f'_{-}\right|$
  and $\left|g'_{+}\right|\le\left|g'_{-}\right|$
  (see, e.g., \citealt{rockafellar:1970}, Theorem 24.1).
  Condition (\ref{eq:tails1}) can be interpreted as saying that
  the tails should shrink fast enough.
  The case $f(t)=g(t)=t^{-1}$ can be considered borderline;
  Theorem \ref{thm:main} is still applicable in this case,
  but it ceases to be applicable for tails
  that shrink less fast.
\end{Remark}

\section{Examples}
\label{sec:examples}

In this section we first define a specific RKHS
and then describe three important games.

\subsection*{Kernels as \Uppercase{s}ource of RKHS}

We start by describing an equivalent language for talking about RKHS.
The \emph{kernel} of an RKHS $\FFF$ on $\mathbf{X}$ is
\begin{equation*}
  \KKK(x,x')
  :=
  \left\langle
    \KKK_x,\KKK_{x'}
  \right\rangle_{\FFF}
\end{equation*}
(equivalently, we could define $\KKK(x,x')$ as $\KKK_x(x')$
or as $\KKK_{x'}(x)$).
There is a simple internal characterization of the kernels $\KKK$ of RKHS.

It is easy to check that the function $\KKK(x,x')$,
as we defined it,
is symmetric
($\KKK(x,x')=\KKK(x',x)$ for all $x,x'\in\mathbf{X}$)
and positive definite
($\sum_{i=1}^m\sum_{j=1}^m \alpha_i\alpha_j \KKK(x_i,x_j)\ge0$
for all $m=1,2,\ldots$, all $(\alpha_1,\ldots,\alpha_m)\in\bbbr^m$,
and all $(x_1,\dots,x_m)\in\mathbf{X}^m$).
On the other hand,
for every symmetric and positive definite $\KKK:\mathbf{X}^2\to\bbbr$
there exists a unique RKHS $\FFF$
such that $\KKK$ is the kernel of $\FFF$
(\citealt{aronszajn:1944}, Th\'eor\`eme 2).

We can see that the notions of a kernel of RKHS
and of a symmetric positive definite function on $\mathbf{X}^2$
have the same content,
and we will sometimes say ``kernel on $\mathbf{X}$''
to mean a symmetric positive definite function
on $\mathbf{X}^2$.
Kernels in this sense are the main source of RKHS in learning theory;
see, e.g., \citet{vapnik:1998}, \citet{scholkopf/smola:2002},
and \citet{shawe-taylor/cristianini:2004}
for numerous examples.
Every kernel on $\mathbf{X}$ is a valid parameter for our decision algorithm;
to apply Theorem \ref{thm:main} we can use the equivalent definition
of $\CC_{\FFF}$,
\begin{equation*}
  \CC_{\FFF}
  :=
  \sup_{x\in\mathbf{X}}
  \sqrt{\KKK(x,x)}.
\end{equation*}

A long list of RKHS together with their kernels is given
\ifJOURNAL by \fi
\ifnotJOURNAL in \fi
\citet{berlinet/thomas-agnan:2004}, \S7.4.
For concreteness, in this section we will use the Sobolev space $\SSS$
of absolutely continuous functions $f$ on $\bbbr$ with finite norm
\begin{equation}\label{eq:Sobolev}
  \left\|f\right\|_{\SSS}
  :=
  \sqrt
  {
    \int_{-\infty}^{\infty}
      f^2(x)
    \D x
    +
    \int_{-\infty}^{\infty}
      (f'(x))^2
    \D x
  };
\end{equation}
its kernel is
\begin{equation*}
  \KKK(x,x')
  =
  \frac{1}{2}
  \exp
  \left(
    -
    \left|
      x-x'
    \right|
  \right)
\end{equation*}
(see \citealt{thomas-agnan:1996}
or \citealt{berlinet/thomas-agnan:2004}, \S7.4, Example 24).
From the last equation we can see that $\CC_{\SSS}=1/\sqrt{2}$.

\subsection*{The \Uppercase{s}quare \Uppercase{l}oss \Uppercase{g}ame}

For the square loss game,
$\Gamma=[0,1]$ and $\lambda(y,\gamma)=(y-\gamma)^2$,
and so we have
\begin{equation}\label{eq:ExpD}
  \Exp_{\lambda}(\gamma)
  =
  \lambda(1,\gamma) - \lambda(0,\gamma)
  =
  (1-\gamma)^2 - \gamma^2
  =
  1 - 2 \gamma,
\end{equation}
\begin{equation*}
  \lambda(p,\gamma)
  =
  p(1-\gamma)^2 + (1-p)\gamma^2
  =
  p(1-p) + (\gamma-p)^2,
\end{equation*}
and
\begin{equation}\label{eq:proper}
  \Gamma_p
  =
  \{p\}.
\end{equation}
Therefore,
\begin{equation*}
  \CC_{\lambda,\FFF}
  =
  \begin{cases}
    \CC_{\FFF}/2 & \text{if $\CC_{\FFF}\ge1$}\\
    (1+\CC^2_{\FFF})/4 & \text{if $\CC_{\FFF}<1$};
  \end{cases}
\end{equation*}
in particular,
$\CC_{\lambda,\SSS}=3/8$ for the Sobolev space (\ref{eq:Sobolev}),
and Theorem~\ref{thm:main} implies
\begin{corollary}\label{cor:square}
  Suppose the decision space is $\mathbf{X}=\bbbr$.
  There is a decision strategy
  that guarantees that, for all $N$ and all decision rules $D\in\SSS$,
  \begin{equation*}
    \sum_{n=1}^N
    (y_n-\gamma_n)^2
    \le
    \sum_{n=1}^N
    (y_n-D(x_n))^2
    +
    \frac{3}{8}
    \left(
      \left\|2D-1\right\|_{\SSS}
      +
      1
    \right)
    \sqrt{N}
  \end{equation*}
  ($2D-1$ is the decision rule
  ``normalized'' to take values in $[-1,1]$).
\end{corollary}

\begin{Remark}\label{rem:illustration}
  The games of this section illustrate Remark \ref{rem:prediction}:
  here the decisions $\gamma_n$ are best interpreted as predictions of $y_n$.
  Loss functions $\lambda$ satisfying (\ref{eq:proper})
  are called \emph{proper scoring rules}.
  Such loss functions ``encourage honesty'':
  it is optimal to predict with the true probability
  (provided it is known).
  We will later see another loss function of this type
  (the log loss function).
\end{Remark}

To illustrate Corollary \ref{cor:square},
suppose there are constants $c>1$ and $d>1$
and a good absolutely continuous decision rule $D:\bbbr\to[0,1]$ such that
$\lvert x_n\rvert\le c$, $n=1,2,\ldots$, and $\lvert D'(x)\rvert\le d$
for all $x\in\mathbf{X}$.
At rounds $N\gg cd^2$ the average loss of our decision algorithm
will be almost as good as (or better than)
the loss of $D$.
We refrain from giving similar illustrations
for the other corollaries in this section.

\subsection*{The \Uppercase{a}bsolute \Uppercase{l}oss \Uppercase{g}ame}

In this game,
$\lambda(y,\gamma)=\lvert y-\gamma\rvert$ with $\Gamma=[0,1]$.
We find:
\begin{equation*}
  \Exp_{\lambda}(\gamma)
  =
  \lambda(1,\gamma) - \lambda(0,\gamma)
  =
  (1-\gamma) - \gamma
  =
  1 - 2 \gamma
\end{equation*}
(the same as in the square loss case,
(\ref{eq:ExpD})),
\begin{equation*}
  \lambda(p,\gamma)
  =
  p(1-\gamma) + (1-p)\gamma
  =
  p + (1-2p)\gamma,
\end{equation*}
and
\begin{equation*}
  \Gamma_p
  =
  \begin{cases}
    \{0\} & \text{if $p<1/2$}\\
    \{1\} & \text{if $p>1/2$}\\
    [0,1] & \text{if $p=1/2$}.
  \end{cases}
\end{equation*}
Therefore,
\begin{equation*}
  \CC_{\lambda,\FFF}
  =
  \frac12
  \sqrt{1+\CC^2_{\FFF}}
\end{equation*}
(in particular, $\CC_{\lambda,\SSS}=\sqrt{6}/4$),
and we have the following corollary of Theorem~\ref{thm:main}.
\begin{corollary}\label{cor:absolute}
  Let $\mathbf{X}=\bbbr$.
  There is a decision strategy that
  produces decisions $\gamma_n$ such that,
  for all $N$ and all $D\in\SSS$,
  \begin{equation}\label{eq:absolute}
    \sum_{n=1}^N
    \left|
      y_n-\gamma_n
    \right|
    \le
    \sum_{n=1}^N
    \left|
      y_n-D(x_n)
    \right|
    +
    \frac{\sqrt{6}}{4}
    \left(
      \left\|2D-1\right\|_{\SSS}
      +
      1
    \right)
    \sqrt{N}.
  \end{equation}
\end{corollary}

\subsection*{The \Uppercase{l}og \Uppercase{l}oss \Uppercase{g}ame}

For the log loss game, $\Gamma=(0,1)$ and
\begin{equation*}
  \lambda(y,\gamma)
  =
  -y\ln\gamma - (1-y)\ln(1-\gamma).
\end{equation*}
For this game,
$\CC_{\lambda,\FFF}<\infty$
(assuming $\CC_{\FFF}<\infty$)
since its tails satisfy
\begin{equation*}
  f'(t)=g'(t)
  =
  -\frac{1}{e^t-1}
  \sim
  -e^{-t}
  =
  O(t^{-2});
\end{equation*}
this will be also clear from the following direct calculation.
Since
\begin{equation*}
  \Exp_{\lambda}(\gamma)
  =
  \lambda(1,\gamma) - \lambda(0,\gamma)
  =
  -\ln\gamma + \ln(1-\gamma)
  =
  \ln\frac{1-\gamma}{\gamma},
\end{equation*}
\begin{equation*}
  \lambda(p,\gamma)
  =
  -p\ln\gamma - (1-p)\ln(1-\gamma)
  =
  \lambda(p,p) + D(p,\gamma)
\end{equation*}
(where $D(p,\gamma):=p\ln\frac{p}{\gamma}+(1-p)\ln\frac{1-p}{1-\gamma}$
is the Kullback distance between $p$ and $\gamma$,
known to take its minimal value in $\gamma$ at $\gamma=p$),
and
$
  \Gamma_p
  =
  \{p\}
$,
we can bound $\CC_{\lambda,\FFF}$ from above as follows:
\begin{multline*}
  \CC^2_{\lambda,\FFF}
  =
  \sup_{p\in(0,1)}
  p(1-p)
  \left(
    \left(
      \ln\frac{1-p}{p}
    \right)^2
    +
    \CC^2_{\FFF}
  \right)
  \ifnotJOURNAL\\\fi
  \le
  \CC^2_{\FFF}/4
  +
  \sup_{p\in(0,1)}
  p(1-p)
  \left(
    \ln\frac{1-p}{p}
  \right)^2\\
  \approx
  \CC^2_{\FFF}/4
  +
  0.439
  \le
  \CC^2_{\FFF}/4
  +
  0.44.
\end{multline*}
Of course,
for specific values of $\CC_{\FFF}$
it is better to find the $\sup_{p\in(0,1)}$ directly,
without using this bound.
Such a direct calculation shows that
$\CC_{\lambda,\SSS}\approx0.693\le0.7$,
and Theorem~\ref{thm:main} now implies the following.
\begin{corollary}\label{cor:log}
  Some decision strategy
  in the log loss game with $\mathbf{X}=\bbbr$
  produces decisions $\gamma_n$ such that,
  for all $N$ and all $D:\mathbf{X}\to(0,1)$
  with the log-likelihood ratio $\ln\frac{D}{1-D}$ in $\SSS$,
  \begin{equation*}
    \sum_{n=1}^N
    \lambda(y_n,\gamma_n)
    \le
    \sum_{n=1}^N
    \lambda(y_n,D(x_n))
    +
    0.7
    \left(
      \left\|\ln\frac{D}{1-D}\right\|_{\SSS}
      +
      1
    \right)
    \sqrt{N}.
  \end{equation*}
\end{corollary}

\section{Idea of the \Uppercase{p}roof of Theorem \ref{thm:main}}
\label{sec:idea}

This section describes the intuition behind the proof.
The following sections,
which carry out the proof,
are formally independent of this section.
We will also describe a general research program
that may lead, it can be hoped, to many other results.

\subsection*{Game-theoretic \Uppercase{p}robability}

Our proof technique is based on a game-theoretic alternative
to the standard measure-theoretic axioms of probability
(\citealt{shafer/vovk:2001}).
Many of the standard laws of probability,
including the weak and strong laws of large numbers,
the central limit theorem,
and the law of the iterated logarithm,
can be restated in terms of perfect information games
involving three key players:
Reality, Forecaster, and Skeptic.
A typical game-theoretic law of probability
states that Skeptic has a strategy
which, without risking bankruptcy,
greatly enriches him if the law is violated.
All such strategies for Skeptic were explicitly constructed
continuous functions;
game-theoretic laws of probability
with a continuous strategy for Skeptic will be called
``continuous laws of probability''.

Game-theoretic probability as developed in \citet{shafer/vovk:2001}
was to a large degree parallel to measure-theoretic probability.
\ifJOURNAL Inspired by \fi\ifnotJOURNAL Following \fi\citet{foster/vohra:1998}
and the literature that this paper spawned,
\ifnotJOURNAL paper \fi\citet{\GTPVIII} pointed out a surprising feature
of game-theoretic probability:
for any continuous law of probability,
Forecaster has a strategy
that prevents Skeptic's capital from growing
(cf.\ Lemma \ref{lem:basic} below).
In other words,
for any continuous law of probability
there is a forecasting strategy
that is perfect as far as this law is concerned
(we will say ``perfect relative to'' this law).
This result was obtained in \citet{\GTPVIII}
for binary forecasting,
and in \citet{\GTPX} it was extended to more general protocols.
Forecasting strategies obtained in this way
from various laws of probability were called ``defensive forecasting'' strategies.

\subsection*{General \Uppercase{p}rocedure}

Now we are ready to describe a general procedure
whose implementation leads,
in the most straightforward case,
to Theorem \ref{thm:main}.

Choose a goal which could be achieved if you knew the true probabilities
generating the observations.
It is important that this goal should be ``practical'',
in the sense of being stated in terms of observable quantities,
such as data, decisions, and observations.
The goal is not allowed to contain theoretical quantities,
such as the true probabilities themselves,
and it should be achievable no matter what the true probabilities are.
Construct a decision strategy which,
using the true probabilities,
leads to the goal.

Realistically, however,
we do not know the true probabilities.
To get rid of them,
isolate the law of probability on which the proof
that your decision strategy achieves the goal depends;
typically, this law can be stated as a continuous game-theoretic law of probability.
(If the proof depends on several laws,
they should first be merged into a single law.)
There is a forecasting strategy whose forecasts
are at least as good as (and often better than)
the true probabilities,
as far as the law you have just isolated is concerned.
It remains to feed your decision strategy with those forecasts.

Implementing this procedure for various interesting goals
appears to be a promising research program.

\subsection*{Introduction to the \Uppercase{p}roof}

In this paper our goal is to achieve (\ref{eq:goal}),
which we roughly rewrite as
\begin{equation*}
  \sum_{n=1}^N
  \lambda(y_n,\gamma_n)
  \lessapprox
  \sum_{n=1}^N
  \lambda(y_n,D(x_n)),
\end{equation*}
where the informal notation $\lessapprox$
is used to mean that the left-hand side does not exceed the right-hand side
plus a quantity small as compared to $N$.
The goal is stated in terms of the observables.

Let us see how our goal could be achieved
if we knew the true probabilities $p_n$ that $y_n=1$
(slightly more formally,
$p_n$ is the conditional probability that $y_n=1$
given the available information).
By the law of large numbers
(see, e.g., \citealt{shiryaev:1996}, Theorem VII.5.4,
for a suitable measure-theoretic statement
and \citealt{shafer/vovk:2001}, Theorem 4.1,
for its game-theoretic counterpart),
we expect
\begin{equation}\label{eq:informalLLN}
  \left|
    \sum_{n=1}^N
    f(p_n,x_n)
    (y_n-p_n)
  \right|
  \ll
  N
\end{equation}
if $f$ is a bounded function
(assumed measurable in the measure-theoretic case).
If $f$ is allowed to range over a function class $\FFF$
that is not excessively wide,
(\ref{eq:informalLLN}) will still continue to hold
uniformly in $f$.

Suppose, for simplicity,
that $\Gamma_p$ is a singleton for all $p\in[0,1]$;
the only element of $\Gamma_p$ will be denoted $G(p)$.
Our decision strategy will make the decision $G(p_n)$ at round $n$,
i.e., the decision that leads to the smallest expected loss.
We will sometimes say that $G$ is our ``choice function''.

Notice that
\begin{equation*}
  \lambda(y,\gamma) - \lambda(p,\gamma)
  =
  (y-p)
  \bigl(
    \lambda(1,\gamma) - \lambda(0,\gamma)
  \bigr)
\end{equation*}
always holds
(this can be checked by subtracting (\ref{eq:expected-loss}) from
$
  \lambda(y,\gamma)
  :=
  y \lambda(1,\gamma)
  +
  (1-y) \lambda(0,\gamma)
$).
In conjunction with the law of large numbers (\ref{eq:informalLLN})
this implies
\begin{multline}\label{eq:chain}
  \sum_{n=1}^N
  \lambda(y_n,\gamma_n)
  =
  \sum_{n=1}^N
  \lambda(y_n,G(p_n))
  \ifnotJOURNAL\\\fi
  =
  \sum_{n=1}^N
  \lambda(p_n,G(p_n))
  +
  \sum_{n=1}^N
  \bigl(
    \lambda(y_n,G(p_n))-\lambda(p_n,G(p_n))
  \bigr)\\
   =
  \sum_{n=1}^N
  \lambda(p_n,G(p_n))
  +
  \sum_{n=1}^N
  (y_n-p_n)
  \bigl(
    \lambda(1,G(p_n))-\lambda(0,G(p_n))
  \bigr)
  \lessapprox
  \sum_{n=1}^N
  \lambda(p_n,G(p_n))\\
  \le
  \sum_{n=1}^N
  \lambda(p_n,D(x_n))
  =
  \sum_{n=1}^N
  \lambda(y_n,D(x_n))
  -
  \sum_{n=1}^N
  \bigl(
    \lambda(y_n,D(x_n))-\lambda(p_n,D(x_n))
  \bigr)\\
  =
  \sum_{n=1}^N
  \lambda(y_n,D(x_n))
  -
  \sum_{n=1}^N
  (y_n-p_n)
  \bigl(
    \lambda(1,D(x_n))-\lambda(0,D(x_n))
  \bigr)
  \ifnotJOURNAL\\\fi
  \lessapprox
  \sum_{n=1}^N
  \lambda(y_n,D(x_n)).
\end{multline}
This shows that we can achieve our goal if we know the true probabilities,
and it remains to replace the true probabilities
with the forecasts that are perfect relative to the law of large numbers.

For clarity,
let us summarize the idea of the proof
expressed by (\ref{eq:chain}).
To show that the actual loss of our decision strategy
does not exceed the actual loss of a decision rule $D$ by much,
we notice that:
\begin{itemize}
\item
  the actual loss
  $\sum_{n=1}^N \lambda(y_n,G(p_n))$
  of our decision strategy is approximately equal,
  by the law of large numbers,
  to the (one-step-ahead conditional) expected loss
  $\sum_{n=1}^N \lambda(p_n,G(p_n))$
  of our strategy;
\item
  since we used the expected loss minimization principle,
  the expected loss of our strategy does not exceed
  the expected loss of $D$;
\item
  the expected loss
  $\sum_{n=1}^N \lambda(p_n,D(x_n))$
  of $D$ is approximately equal
  to its actual loss
  $\sum_{n=1}^N \lambda(y_n,D(x_n))$
  (by the law of large numbers).
\end{itemize}

To get the strongest possible result,
we will have to use more specific laws of probability
than the general law of large numbers.
It will be convenient to use the following informal terminology
introduced in \citet{\GTPXIII}.
Let $p_n$ be the forecasts output by some forecasting strategy
(rather than the true probabilities).
We say that the forecasting strategy has good \emph{calibration-cum-resolution}
if the left-hand side of (\ref{eq:informalLLN}) is much less than $N$
for a relatively wide class of functions $f:[0,1]\times\mathbf{X}\to\bbbr$ and large $N$.
We say that the strategy has good \emph{calibration}
if
\begin{equation*}
  \left|
    \sum_{n=1}^N
    (y_n-p_n)
    f(p_n)
  \right|
  \ll
  N
\end{equation*}
for a wide class of functions $f:[0,1]\to\bbbr$ and large $N$.
Finally,
we say that the strategy has good \emph{resolution}
if
\begin{equation*}
  \left|
    \sum_{n=1}^N
    (y_n-p_n)
    f(x_n)
  \right|
  \ll
  N
\end{equation*}
for a wide class of $f:\mathbf{X}\to\bbbr$ and for large $N$.
For a detailed discussion and examples,
see \citealt{\GTPXIII}.

Notice that in applying the law of large numbers
to establishing the two approximate inequalities in (\ref{eq:chain})
we need not general $f=f(p,x)$
but only $f=f(p)$ (known in advance) and $f=f(x)$.
In particular, we only need calibration and resolution separately,
not calibration-cum-resolution.
These are the two specific probability laws we will be concerned with.

The requirement that $\Gamma_p$ should always be a singleton
(in fact, we will even need the function $G(p)$ to be continuous)
is restrictive:
for example, it is not satisfied for the absolute loss function.
To deal with this problem,
we will have to consider forecasting strategies
that output extended forecasts $(p_n,q_n)\in[0,1]^2$,
where $p_n$ is the forecast of $y_n$
and the extra component $q_n$ will play a more technical role.

The next section is devoted to constructing a perfect forecasting strategy
relative to the law of large numbers.
In the following section we will be able to prove Theorem \ref{thm:main}.

\section{The \Uppercase{a}lgorithm of \Uppercase{l}arge \Uppercase{n}umbers}
\label{sec:ALN}

This section is the core of our proof of Theorem \ref{thm:main}.
First we describe a forecasting protocol
in which Forecaster tries to predict the observations chosen by Reality.
Following \citet{shafer/vovk:2001},
we introduce another player, Skeptic,
who is allowed to bet
at the odds implied by Forecaster's moves.

\bigskip

\noindent
\textsc{Binary Forecasting Game I}

\noindent
\textbf{Players:} Reality, Forecaster, Skeptic

\noindent
\textbf{Protocol:}

\parshape=7
\IndentI   \WidthI
\IndentII  \WidthII
\IndentII  \WidthII
\IndentII  \WidthII
\IndentII  \WidthII
\IndentII  \WidthII
\IndentI   \WidthI
\noindent
FOR $n=1,2,\dots$:\\
  Reality announces $x_n\in\mathbf{X}$.\\
  Forecaster announces $(p_n,q_n)\in[0,1]^2$.\\
  Skeptic announces $s_n\in\bbbr$.\\
  Reality announces $y_n\in\{0,1\}$.\\
  $\K_n := \K_{n-1} + s_n (y_n-p_n)$.\\
END FOR.

\bigskip

\noindent
The real forecast is $p_n$ (the ``probability'' that $y_n=1$),
which is interpreted as the price
Forecaster charges for a ticket paying $y_n$;
$s_n$ is the number of tickets Skeptic decides to buy.
The protocol describes not only the players' moves
but also the changes in Skeptic's capital $\K_n$;
its initial value $\K_0$ can be an arbitrary real number.
Skeptic demonstrates that the forecasts are poor
if he manages to multiply his initial capital (assumed positive) manyfold
without risking bankruptcy
(i.e., $\K_n$ becoming negative).
Forecaster also provides an additional number $q_n\in[0,1]$
which does not affect Skeptic's capital;
intuitively,
the role of $q_n$ is to help those of Forecaster's customers
who find themselves in a position of Buridan's ass
(find two or more actions equally attractive
in view of the forecast $p_n$)
to break the tie.

The main difference between our decision protocol
(stated at the beginning of \S\ref{sec:main})
and the protocols of this section
is that in the latter
Forecaster implicitly claims
(by pricing the tickets)
that he has the fullest possible knowledge of the way Reality chooses the observations,
and Skeptic tries to prove him wrong by gambling against him.
In the decision protocol,
Decision Maker does no make any such claims
and simply tries to minimize his losses.

It will be convenient to make the set $[0,1]^2$
from which the forecasts $(p_n,q_n)$ are chosen
into a topological space.
The \emph{lexicographic square} $\pounds$ is defined to be the set $[0,1]^2$
equipped with the following linear order:
if $(x_1,y_1)$ and $(x_2,y_2)$ are two points in $\pounds$,
$(x_1,y_1)<(x_2,y_2)$ means that either $x_1<x_2$
or $x_1=x_2,y_1<y_2$.
(Cf.\ \citealt{engelking:1989local}, Problem 3.12.3(d).)
The topology on the lexicographic square is, as usual,
generated by the open intervals
\begin{equation*}
  (a,b)
  :=
  \left\{
    u\in\pounds
    \st
    a<u<b
  \right\}
  ,
\end{equation*}
$a$ and $b$ ranging over $\pounds$.
As a topological space,
the lexicographic square is 
normal
(\citealt{engelking:1989local}, Problem 1.7.4(d)),
compact
(\citealt{engelking:1989local}, Problem 3.12.3(a),
\citealt{kelley:1957}, Problem 5.C),
and connected
(\citealt{engelking:1989local}, Problem 6.3.2(a),
\citealt{kelley:1957}, Problem 1.I(d)).

As in \citet{\GTPVIII},
we will see that for any continuous strategy for Skeptic
there exists a strategy for Forecaster
that does not allow Skeptic's capital to grow,
regardless of what Reality is doing.
To state this observation in its strongest form,
we make Skeptic announce his strategy
for each round
before Forecaster's move on that round
rather than announce his full strategy at the beginning of the game.
Therefore,
we consider the following perfect-information game:

\bigskip

\noindent
\textsc{Binary Forecasting Game II}

\noindent
\textbf{Players:} Reality, Forecaster, Skeptic

\noindent
\textbf{Protocol:}

\parshape=7
\IndentI   \WidthI
\IndentII  \WidthII
\IndentII  \WidthII
\IndentII  \WidthII
\IndentII  \WidthII
\IndentII  \WidthII
\IndentI   \WidthI
\noindent
FOR $n=1,2,\dots$:\\
  Reality announces $x_n\in\mathbf{X}$.\\
  Skeptic announces continuous $S_n:\pounds\to\bbbr$.\\
  Forecaster announces $(p_n,q_n)\in\pounds$.\\
  Reality announces $y_n\in\{0,1\}$.\\
  $\K_n := \K_{n-1} + S_n(p_n,q_n) (y_n-p_n)$.\\
END FOR.

\bigskip

\begin{lemma}\label{lem:basic}
  Forecaster has a strategy in Binary Forecasting Game II
  that ensures $\K_0\ge\K_1\ge\K_2\ge\cdots$.
\end{lemma}

Before proving this lemma, we will need another lemma,
which will play the role of the Intermediate Value Theorem,
used in \citet{\GTPXIII}.
\begin{lemma}\label{lem:intermediate}
  If a continuous function $f:\pounds\to\bbbr$ takes both positive and negative values,
  there exists $x\in\pounds$ such that $f(x)=0$.
\end{lemma}
\begin{proof}
  A continuous image of a connected compact set
  is connected
  (\citealt{engelking:1989local}, Theorem 6.1.4)
  and compact
  (\citealt{engelking:1989local}, Theorem 3.1.10).
  Therefore, $f(\pounds)$ is a closed interval.
  \qedtext
\end{proof}

\begin{Proof}{of Lemma \ref{lem:basic}}
  Forecaster can now use the following strategy to ensure $\K_0\ge\K_1\ge\cdots$:
  \begin{itemize}
  \item
    if the function $S_n(p,q)$ takes value 0,
    choose $(p_n,q_n)$ such that $S_n(p_n,q_n)=0$;
  \item
    if $S_n$ is always positive,
    take $p_n:=1$ and choose $q_n\in[0,1]$ arbitrarily;
  \item
    if $S_n$ is always negative,
    take $p_n:=0$ and choose $q_n\in[0,1]$ arbitrarily.
    \qedtext
  \end{itemize}
\end{Proof}

A kernel $\KKK$ on $\pounds\times\mathbf{X}$ is \emph{forecast-continuous}
if the function $\KKK((p,q,x),(p',q',x'))$ is continuous in $(p,q,p',q')\in\pounds^2$,
for each fixed $(x,x')\in\mathbf{X}^2$.
(Kernels on $\pounds\times\mathbf{X}$ are defined analogously to kernels on $\mathbf{X}$.)
For such a kernel the function
\begin{equation}\label{eq:function}
  S_n(p,q)
  :=
  \sum_{i=1}^{n-1} \KKK((p,q,x_n),(p_i,q_i,x_i))(y_i-p_i)
  +
  \frac12
  \KKK((p,q,x_n),(p,q,x_n))(1-2p)
\end{equation}
is continuous in $(p,q)\in\pounds$.

\bigskip

\noindent
\textsc{The lexicographic algorithm of large numbers ($\pounds$ALN)}

\noindent
\textbf{Parameter:} forecast-continuous kernel $\KKK$ on $\pounds\times\mathbf{X}$

\parshape=8
\IndentI   \WidthI
\IndentII  \WidthII
\IndentII  \WidthII
\IndentII  \WidthII
\IndentIII \WidthIII
\IndentIII \WidthIII
\IndentII  \WidthII
\IndentI   \WidthI
\noindent
FOR $n=1,2,\dots$:\\
  Read $x_n\in\mathbf{X}$.\\
  Define $S_n(p,q)$ by (\ref{eq:function}), $(p,q)\in\pounds$.\\
  Output any root $(p,q)$ of $S_n(p,q)=0$ as $(p_n,q_n)$;\\
    if there are no roots,\\
    set $p_n:=(1+\sign S_n)/2$ and set $q_n$ to any number in $[0,1]$.\\
  Read $y_n\in\{0,1\}$.\\
END FOR.

\bigskip

\noindent
(Notice that $\sign S_n$ is well defined by Lemma \ref{lem:intermediate}.)
It is well known that
there exists a function $\Phi:\pounds\times\mathbf{X}\to\HHH$
(a \emph{feature mapping} taking values in a Hilbert space $\HHH$)
such that
\begin{equation}\label{eq:K}
  \KKK(a,b)
  =
  \Phi(a)\cdot\Phi(b),
  \enspace
  \forall a,b\in\pounds\times\mathbf{X}.
\end{equation}
(For example,
we can take the RKHS on $\pounds\times\mathbf{X}$ with kernel $\KKK$ as $\HHH$
and take $a\mapsto\KKK_a$ as the feature mapping $\Phi$;
there are, however,
easier and more transparent constructions.)
It can be shown that $\Phi(p,q,x)$ is \emph{forecast-continuous},
i.e., continuous in $(p,q)\in\pounds$ for each fixed $x\in\mathbf{X}$,
if and only if the kernel $\KKK$ defined by~(\ref{eq:K}) is forecast-continuous
(see, e.g., \citealt{\GTPXIII}, Appendix~B).
\begin{theorem}\label{thm:K29}
  Let $\KKK$ be the kernel
  defined by~(\ref{eq:K})
  for a forecast-continuous feature mapping $\Phi:\pounds\times\mathbf{X}\to\HHH$.
  The lexicographic algorithm of large numbers with parameter $\KKK$ outputs $(p_n,q_n)$
  such that
  \begin{equation}\label{eq:K29}
    \left\|
      \sum_{n=1}^N
      (y_n-p_n)
      \Phi(p_n,q_n,x_n)
    \right\|^2
    \le
    \sum_{n=1}^N
    p_n(1-p_n)
    \left\|
      \Phi(p_n,q_n,x_n)
    \right\|^2
  \end{equation}
  always holds for all $N=1,2,\dots$.
\end{theorem}
\begin{proof}
  Following $\pounds$ALN
  Forecaster ensures that Skeptic will never increase his capital
  with the strategy
  \begin{multline}\label{eq:strategy}
    s_n
    :=
    \sum_{i=1}^{n-1}
    \KKK
    \left(
      (p_n,q_n,x_n),(p_i,q_i,x_i)
    \right)
    (y_i-p_i)\\
    +
    \frac12
    \KKK
    \left(
      (p_n,q_n,x_n),(p_n,q_n,x_n)
    \right)
    (1-2p_n).
  \end{multline}
  Using the formula
  \begin{equation*}
    (y_n-p_n)^2
    =
    p_n(1-p_n)
    +
    (1-2p_n)
    (y_n-p_n)
  \end{equation*}
  (which can be checked by setting $y_n:=0$ and $y_n:=1$),
  we can see that the increase in Skeptic's capital
  when he follows~(\ref{eq:strategy})
  is
  \begin{align*}
    \K_N-\K_0
    &=
    \sum_{n=1}^N
    s_n(y_n-p_n)\\
    &=
    \sum_{n=1}^N
    \sum_{i=1}^{n-1}
    \KKK
    \left(
      (p_n,q_n,x_n),(p_i,q_i,x_i)
    \right)
    (y_n-p_n)
    (y_i-p_i)\\
    &\quad{}+
    \frac12
    \sum_{n=1}^N
    \KKK
    \left(
      (p_n,q_n,x_n),(p_n,q_n,x_n)
    \right)
    (1-2p_n)
    (y_n-p_n)\\
    &=
    \frac12
    \sum_{n=1}^N
    \sum_{i=1}^N
    \KKK
    \left(
      (p_n,q_n,x_n),(p_i,q_i,x_i)
    \right)
    (y_n-p_n)
    (y_i-p_i)\\
    &\quad{}-
    \frac12
    \sum_{n=1}^N
    \KKK
    \left(
      (p_n,q_n,x_n),(p_n,q_n,x_n)
    \right)
    (y_n-p_n)^2\\
    &\quad{}+
    \frac12
    \sum_{n=1}^N
    \KKK
    \left(
      (p_n,q_n,x_n),(p_n,q_n,x_n)
    \right)
    (1-2p_n)
    (y_n-p_n)\\
    &=
    \frac12
    \sum_{n=1}^N
    \sum_{i=1}^N
    \KKK
    \left(
      (p_n,q_n,x_n),(p_i,q_i,x_i)
    \right)
    (y_n-p_n)
    (y_i-p_i)\\
    &\quad{}-
    \frac12
    \sum_{n=1}^N
    \KKK
    \left(
      (p_n,q_n,x_n),(p_n,q_n,x_n)
    \right)
    p_n(1-p_n)\\
    &=
    \frac12
    \left\|
      \sum_{n=1}^N
      (y_n-p_n)
      \Phi(p_n,q_n,x_n)
    \right\|^2
    -
    \frac12
    \sum_{n=1}^N
    p_n(1-p_n)
    \left\|
      \Phi(p_n,q_n,x_n)
    \right\|^2,
  \end{align*}
  which immediately implies~(\ref{eq:K29}).
  \qedtext
\end{proof}

\subsection*{Resolution}

This subsection makes the next step in our proof of Theorem~\ref{thm:main}.
Its forecasting protocol is:

\bigskip

\parshape=5
\IndentI  \WidthI
\IndentII \WidthII
\IndentII \WidthII
\IndentII \WidthII
\IndentI  \WidthI
\noindent
FOR $n=1,2,\dots$:\\
  Reality announces $x_n\in\mathbf{X}$.\\
  Forecaster announces $(p_n,q_n)\in[0,1]$.\\
  Reality announces $y_n\in\{0,1\}$.\\
END FOR.

\bigskip

\noindent
Our goal is to prove the following result
(although in \S\ref{sec:proof} we will need a slight modification of this result
rather than the result itself).
\begin{theorem}\label{thm:resolution}
  Let $\FFF$ be an RKHS on $\mathbf{X}$.
  The forecasts $(p_n,q_n)$ output by $\pounds$ALN always satisfy
  \begin{equation*}
    \left|
      \sum_{n=1}^N
      (y_n-p_n)
      f(x_n)
    \right|
    \le
    \frac{\CC_{\FFF}}{2}
    \left\|f\right\|_{\FFF}
    \sqrt{N}
  \end{equation*}
  for all $N$ and all functions $f\in\FFF$.
\end{theorem}

\begin{proof}
Applying ALN to the feature mapping
$
  x\in\mathbf{X}
  \mapsto
  \KKK_x\in\FFF
$
and using (\ref{eq:K29}),
we obtain
\begin{multline}\label{eq:simple-resolution}
  \left|
    \sum_{n=1}^N
    (y_n-p_n)
    f(x_n)
  \right|
  =
  \left|
    \sum_{n=1}^N
    (y_n-p_n)
    \left\langle
      \KKK_{x_n}, f
    \right\rangle_{\FFF}
  \right|\\
  =
  \left|
    \left\langle
      \sum_{n=1}^N
      (y_n-p_n)
      \KKK_{x_n},
      f
    \right\rangle_{\FFF}
  \right|
  \le
  \left\|
    \sum_{n=1}^N
    (y_n-p_n)
    \KKK_{x_n}
  \right\|_{\FFF}
  \left\|
    f
  \right\|_{\FFF}\\
  \le
  \left\|
    f
  \right\|_{\FFF}
  \sqrt
  {
    \sum_{n=1}^N
    p_n(1-p_n)
    \KKK(x_n,x_n)
  }
  \le
  \CC_{\FFF}
  \left\|
    f
  \right\|_{\FFF}
  \sqrt
  {
    \sum_{n=1}^N
    p_n(1-p_n)
  }
\end{multline}
for any $f\in\FFF$.
\qedtext
\end{proof}

\begin{Remark}
  In the terminology introduced in the previous section,
  Theorem \ref{thm:resolution} is about resolution.
  This is sufficient for the purpose of this paper,
  but it is easy to see that similar statements hold
  for calibration-cum-resolution and calibration.
  For example, let $\FFF$ be an RKHS on $\pounds\times\mathbf{X}$.
  The forecasts $(p_n,q_n)$ output by $\pounds$ALN always satisfy
  \begin{equation*}
    \left|
      \sum_{n=1}^N
      (y_n-p_n)
      f(p_n,q_n,x_n)
    \right|
    \le
    \frac{\CC_{\FFF}}{2}
    \left\|f\right\|_{\FFF}
    \sqrt{N}
  \end{equation*}
  for all $N$ and all functions $f\in\FFF$.
\end{Remark}

\section{Proof of Theorem \ref{thm:main}}
\label{sec:proof}

Before starting the proof proper,
we need to discuss two topics:
choosing a suitable choice function
and ``mixing'' different feature mappings.

\subsection*{The \Uppercase{c}anonical \Uppercase{c}hoice \Uppercase{f}unction}

Let us say that a straight line
$(1-p)x+py=c$ in the $(x,y)$-plane,
where $p\in[0,1]$ and $c\in\bbbr$,
is \emph{southwest of} the superdecision set $\Sigma$
(defined by (\ref{eq:super}))
if
\begin{equation*}
  \forall(x,y)\in\Sigma:
  (1-p)x+py\ge c.
\end{equation*}
For each $p\in[0,1]$
let $c(p)$ be the largest $c$
(which obviously exists)
such that the line
$(1-p)x+py=c$ is southwest of $\Sigma$.
It is clear that, for $p\in(0,1)$,
the line $(1-p)x+py=c(p)$
intersects $\Sigma$ and the intersection,
being compact and convex,
has the form $[A(p),B(p)]$,
where $A(p)$ and $B(p)$ are points
(perhaps $A(p)=B(p)$)
on the line.
For concreteness,
let $A(p)$ be northwest of $B(p)$
(i.e.,
if $A(p)=(A_0,A_1)$ and $B(p)=(B_0,B_1)$,
we assume that $A_0\le B_0$ and $A_1\ge B_1$).
Now we can define the \emph{canonical choice function} $G$
associated with $(\Gamma,\lambda)$ as follows:
\begin{itemize}
\item
  if $0<p<1$ and $q\in[0,1]$,
  $G(p,q)$ is defined to be any $\gamma\in\Gamma$ satisfying
  \begin{equation*}
    (\lambda(0,\gamma),\lambda(1,\gamma))
    =
    (1-q)A(p) + qB(p);
  \end{equation*}
  the existence of such a $\gamma$ is obvious;
\item
  if $p=0$ and $q\in[0,1]$,
  $G(p,q)$ is defined to be any fixed $\gamma_0\in\Gamma$ satisfying
  \begin{equation*}
    (\lambda(0,\gamma_0),\lambda(1,\gamma_0))
    =
    \left(
      C_0,f(C_0)
    \right)
  \end{equation*}
  ($C_0$ and $f$ are defined in (\ref{eq:C-01}) and (\ref{eq:Eastern}));
  if $f(C_0)=\infty$,
  such a $\gamma_0$ does not exist and $G(p,q)$ is undefined;
\item
  if $p=1$ and $q\in[0,1]$,
  $G(p,q)$ is defined to be any fixed $\gamma_1\in\Gamma$ such that
  \begin{equation*}
    (\lambda(0,\gamma_1),\lambda(1,\gamma_1))
    =
    \left(
      g(C_1),C_1
    \right)
  \end{equation*}
  ($C_1$ and $g$ are defined in (\ref{eq:C-01}) and (\ref{eq:Northern}));
  if $g(C_1)=\infty$,
  such a $\gamma_1$ does not exist and $G(p,q)$ is undefined.
\end{itemize}
It is easy to see that the function
$(\lambda(0,G(p,q)),\lambda(1,G(p,q)))$
is continuous in $(p,q)\in\dom G$
and, therefore,
$\Exp_{\lambda,G}(p,q):=\Exp_{\lambda}(G(p,q))$ is continuous in $(p,q)\in\dom G$.
We defined $G$ in such a way that it is a ``perfect'' choice function:
$\lambda(p,G(p,q))=\inf_{\gamma\in\Gamma}\lambda(p,\gamma)$
for virtually all $(p,q)$
(in any case, for all $(p,q)\in\dom G$).

\subsection*{Mixing}

In the proof of Theorem \ref{thm:main}
we will mix the feature mapping $\Phi_0(p,q,x):=\Exp_{\lambda,G}(p,q)$
(into $\HHH_0:=\bbbr$)
and the feature mapping $\Phi_1(p,q,x):=\KKK_x$
used in the proof of Theorem \ref{thm:resolution}
(as discussed in \S\ref{sec:idea},
we will have to achieve two goals simultaneously,
only one of them connected with resolution).
This can be done using the following corollary
of Theorem \ref{thm:K29}.
\begin{corollary}\label{cor:mixture}
  Let $\Phi_j:\pounds\times\mathbf{X}\to\HHH_j$, $j=0,1$,
  be forecast-continuous mappings from $\pounds\times\mathbf{X}$
  to Hilbert spaces $\HHH_j$.
  The forecasts output by $\pounds$ALN with a suitable kernel parameter always satisfy
  \ifJOURNAL
  \begin{equation*}
    \left\|
      \sum_{n=1}^N
      (y_n-p_n)
      \Phi_j(p_n,q_n,x_n)
    \right\|^2_{\HHH_j}
    \le
    \sum_{n=1}^N
    p_n(1-p_n)
    \left(
      \left\|
        \Phi_0(p_n,q_n,x_n)
      \right\|^2_{\HHH_0}
      +
      \left\|
        \Phi_1(p_n,q_n,x_n)
      \right\|^2_{\HHH_1}
    \right)
  \end{equation*}
  \fi
  \ifnotJOURNAL
  \begin{multline*}
    \left\|
      \sum_{n=1}^N
      (y_n-p_n)
      \Phi_j(p_n,q_n,x_n)
    \right\|^2_{\HHH_j}\\
    \le
    \sum_{n=1}^N
    p_n(1-p_n)
    \left(
      \left\|
        \Phi_0(p_n,q_n,x_n)
      \right\|^2_{\HHH_0}
      +
      \left\|
        \Phi_1(p_n,q_n,x_n)
      \right\|^2_{\HHH_1}
    \right)
  \end{multline*}
  \fi
  for all $N$ and for both $j=0$ and $j=1$.
\end{corollary}
\begin{proof}
  Define the direct sum $\HHH$ of $\HHH_0$ and $\HHH_1$
  as the Cartesian product $\HHH_0\times\HHH_1$
  equipped with the inner product
  \begin{equation*}
    \langle g, g' \rangle_{\HHH}
    =
    \left\langle
      (g_0,g_1),
      (g'_0,g'_1)
    \right\rangle_{\HHH}
    :=
    \sum_{j=0}^{1}
    \langle g_j, g'_j\rangle_{\HHH_j}.
  \end{equation*}
  Now we can define $\Phi:\pounds\times\mathbf{X}\to\HHH$ by
  \begin{equation*}
    \Phi(p,q,x)
    :=
    \left(
      \Phi_0(p,q,x),
      \Phi_1(p,q,x)
    \right);
  \end{equation*}
  the corresponding kernel is
  \begin{multline*}
    \KKK((p,q,x),(p',q',x'))
    :=
    \left\langle
      \Phi(p,q,x),
      \Phi(p',q',x')
    \right\rangle_{\HHH}\\
    =
    \sum_{j=0}^{1}
    \left\langle
      \Phi_j(p,q,x),
      \Phi_j(p',q',x')
    \right\rangle_{\HHH_j}
    =
    \sum_{j=0}^{1}
    \KKK_j((p,q,x),(p',q',x')),
  \end{multline*}
  where $\KKK_0$ and $\KKK_1$ are the kernels corresponding to $\Phi_0$ and $\Phi_1$,
  respectively.
  It is clear that this kernel is forecast-continuous.
  Applying $\pounds$ALN to it and using~(\ref{eq:K29}),
  we obtain
  \begin{multline*}
    \left\|
      \sum_{n=1}^N
      (y_n-p_n)
      \Phi_j(p_n,q_n,x_n)
    \right\|^2_{\HHH_j}\\
    \le
    \left\|
      \left(
        \sum_{n=1}^N
        (y_n-p_n)
        \Phi_0(p_n,q_n,x_n),
        \sum_{n=1}^N
        (y_n-p_n)
        \Phi_1(p_n,q_n,x_n)
      \right)
    \right\|^2_{\HHH}\\
    =
    \left\|
      \sum_{n=1}^N
      (y_n-p_n)
      \Phi(p_n,q_n,x_n)
    \right\|^2_{\HHH}
    \le
    \sum_{n=1}^N
    p_n(1-p_n)
    \left\|
      \Phi(p_n,q_n,x_n)
    \right\|^2_{\HHH}\\
    =
    \sum_{n=1}^N
    p_n(1-p_n)
    \sum_{j=0}^{1}
    \left\|
      \Phi_j(p_n,q_n,x_n)
    \right\|^2_{\HHH_j}.
    \qedmath
  \end{multline*}
\end{proof}

Merging $\Phi_0$ and $\Phi_1$ by Corollary \ref{cor:mixture},
we obtain
\begin{multline}\label{eq:calibration}
  \left|
    \sum_{n=1}^N
    (y_n-p_n)
    \Exp_{\lambda,G}(p_n,q_n)
  \right|
  =
  \left\|
    \sum_{n=1}^N
    (y_n-p_n)
    \Phi_0(p_n,q_n,x_n)
  \right\|_{\bbbr}\\
  \le
  \sqrt
  {
    \sum_{n=1}^N
    p_n(1-p_n)
    \left(
      \Exp^2_{\lambda,G}(p_n,q_n)
      +
      \KKK(x_n,x_n)
    \right)
  }
\end{multline}
and,
using (\ref{eq:simple-resolution}),
\begin{multline}\label{eq:resolution}
  \left|
    \sum_{n=1}^N
    (y_n-p_n)
    f(x_n)
  \right|
  \le
  \left\|
    \sum_{n=1}^N
    (y_n-p_n)
    \KKK_{x_n}
  \right\|_{\FFF}
  \left\|
    f
  \right\|_{\FFF}\\
  =
  \left\|
    \sum_{n=1}^N
    (y_n-p_n)
    \Phi_1(p_n,q_n,x_n)
  \right\|_{\FFF}
  \left\|
    f
  \right\|_{\FFF}\\
  \le
  \left\|
    f
  \right\|_{\FFF}
  \sqrt
  {
    \sum_{n=1}^N
    p_n(1-p_n)
    \left(
      \Exp_{\lambda,G}^2(p_n,q_n)
      +
      \KKK(x_n,x_n)
    \right)
  },
\end{multline}
for each function $f\in\FFF$.

\subsection*{Proof: Part I}

In this subsection we will assume that $\dom G=\pounds$.
Subtracting (\ref{eq:expected-loss}) from
$
  \lambda(y,\gamma)
  =
  y \lambda(1,\gamma)
  +
  (1-y) \lambda(0,\gamma)
$,
we obtain
\begin{equation}\label{eq:fundamental}
  \lambda(y,\gamma) - \lambda(p,\gamma)
  =
  (y-p)
  \bigl(
    \lambda(1,\gamma) - \lambda(0,\gamma)
  \bigr)
  =
  (y-p)
  \Exp_{\lambda}(\gamma)
  \ifJOURNAL.\fi
\end{equation}
\ifnotJOURNAL
  (we already did this in \S\ref{sec:idea},
  but we promised that the rest of the paper
  would be formally independent of \S\ref{sec:idea}).
\fi
Using the last equality and (\ref{eq:calibration})--(\ref{eq:resolution}),
we obtain for the decision strategy $\gamma_n:=G(p_n,q_n)$
based on the $(p_n,q_n)$ output by $\pounds$ALN with the merged kernel as parameter:
\begin{align*}
  &\sum_{n=1}^N
  \lambda(y_n,\gamma_n)
  =
  \sum_{n=1}^N
  \lambda(y_n,G(p_n,q_n))\\
  &=
  \sum_{n=1}^N
  \lambda(p_n,G(p_n,q_n))
  +
  \sum_{n=1}^N
  \bigl(
    \lambda(y_n,G(p_n,q_n))-\lambda(p_n,G(p_n,q_n))
  \bigr)\\
  &=
  \sum_{n=1}^N
  \lambda(p_n,G(p_n,q_n))
  +
  \sum_{n=1}^N
  (y_n-p_n)
  \Exp_{\lambda,G}(p_n,q_n)\\
  &\le
  \sum_{n=1}^N
  \lambda(p_n,G(p_n,q_n))
  +
  \sqrt
  {
    \sum_{n=1}^N
    p_n(1-p_n)
    \left(
      \Exp_{\lambda,G}^2(p_n,q_n)+\KKK(x_n,x_n)
    \right)
  }\\
  &\le
  \sum_{n=1}^N
  \lambda(p_n,D(x_n))
  +
  \sqrt
  {
    \sum_{n=1}^N
    p_n(1-p_n)
    \left(
      \Exp_{\lambda,G}^2(p_n,q_n)+\KKK(x_n,x_n)
    \right)
  }\\
  &=
  \sum_{n=1}^N
  \lambda(y_n,D(x_n))
  -
  \sum_{n=1}^N
  \bigl(
    \lambda(y_n,D(x_n))-\lambda(p_n,D(x_n))
  \bigr)\\
  &\quad{}+
  \sqrt
  {
    \sum_{n=1}^N
    p_n(1-p_n)
    \left(
      \Exp_{\lambda,G}^2(p_n,q_n)+\KKK(x_n,x_n)
    \right)
  }\\
  &=
  \sum_{n=1}^N
  \lambda(y_n,D(x_n))
  -
  \sum_{n=1}^N
  (y_n-p_n)
  \Exp_{\lambda,D}(x_n)
  \bigr)\\
  &\quad{}+
  \sqrt
  {
    \sum_{n=1}^N
    p_n(1-p_n)
    \left(
      \Exp_{\lambda,G}^2(p_n,q_n)+\KKK(x_n,x_n)
    \right)
  }\\
  &\le
  \sum_{n=1}^N
  \lambda(y_n,D(x_n))
  +
  \left\|\Exp_{\lambda,D}\right\|_{\FFF}
  \sqrt
  {
    \sum_{n=1}^N
    p_n(1-p_n)
    \left(
      \Exp_{\lambda,G}^2(p_n,q_n)+\KKK(x_n,x_n)
    \right)
  }\\
  &\quad{}+
  \sqrt
  {
    \sum_{n=1}^N
    p_n(1-p_n)
    \left(
      \Exp_{\lambda,G}^2(p_n,q_n)+\KKK(x_n,x_n)
    \right)
  }\\
  &=
  \sum_{n=1}^N
  \lambda(y_n,D(x_n))
  +
  \left(
    \left\|\Exp_{\lambda,D}\right\|_{\FFF} + 1
  \right)
  \sqrt
  {
    \sum_{n=1}^N
    p_n(1-p_n)
    \left(
      \Exp_{\lambda,G}^2(p_n,q_n)+\KKK(x_n,x_n)
    \right)
  }\\
  &\le
  \sum_{n=1}^N
  \lambda(y_n,D(x_n))
  +
  \left(
    \left\|\Exp_{\lambda,D}\right\|_{\FFF} + 1
  \right)
  \CC_{\lambda,\FFF}
  \sqrt{N}.
\end{align*}

It remains to show that $\CC_{\lambda,\FFF}<\infty$
(assuming $\CC_{\FFF}<\infty$,
here and in the rest of this section).
In this case, $\dom G=\pounds$,
this is easy:
essentially, this is the case of a bounded loss function
(the reservation ``essentially'' is needed
since $\Gamma$ can contain ``litter''---%
decisions dominated by other decisions in $\Gamma$).
Since $\Exp_{\lambda,G}$ is continuous and $\pounds$ is compact,
\begin{equation*}
  \sup_{(p,q)\in\pounds}
  p(1-p)
  \left(
    \Exp_{\lambda,G}^2(p,q)
    +
    \CC^2_{\FFF}
  \right)
  <
  \infty.
\end{equation*}

\subsection*{Proof: Part II}

The \emph{stripped lexicographic square} is the subset
\begin{equation*}
  \starpoundsstar
  :=
  (0,1) \times [0,1]
\end{equation*}
of $\pounds$.
In this subsection we consider the case $\dom G=\starpoundsstar$.

The order and topology on $\starpoundsstar$ are inherited from $\pounds$.
The following analogue of Lemma \ref{lem:intermediate}
still holds.
\begin{lemma}
  If a continuous function $f:\starpoundsstar\to\bbbr$
  takes both positive and negative values,
  it also takes the value $0$.
\end{lemma}
\begin{proof}
  See the proof of Lemma \ref{lem:intermediate};
  $f(\starpoundsstar)$ is still a connected set in $\bbbr$.
  \qedtext
\end{proof}

A kernel $\KKK$ on $\starpoundsstar\times\mathbf{X}$ is \emph{forecast-continuous}
if the function
$
  \KKK
  \left(
    (p,q,x),
    (p',q',x')
  \right)
$
is continuous in $(p,q,p',q')\in(\starpoundsstar)^2$.
The function (\ref{eq:function}) is then continuous in $(p,q)\in\starpoundsstar$,
and for our current kernel
\begin{equation}\label{eq:kernel-theorem}
  \KKK
  \left(
    (p,q,x),
    (p',q',x')
  \right)
  =
  \Exp_{\lambda,G}(p,q)
  \Exp_{\lambda,G}(p',q')
  +
  \left\langle
    \KKK_x,
    \KKK_{x'}
  \right\rangle_{\FFF}
\end{equation}
it equals
\begin{multline}\label{eq:kernel-chain}
  S_n(p,q)
  =
  \sum_{i=1}^{n-1}
  \Bigl(
    \Exp_{\lambda,G}(p,q)
    \Exp_{\lambda,G}(p_i,q_i)
    +
    \left\langle
      \KKK_{x_n},
      \KKK_{x_i}
    \right\rangle_{\FFF}
  \Bigr)
  (y_i-p_i)
  \\+
  \frac12
  \Bigl(
    \Exp_{\lambda,G}^2(p,q)
    +
    \left\|
      \KKK_{x_n}
    \right\|^2_{\FFF}
  \Bigr)
  (1-2p)
  \\=
  A \Exp_{\lambda,G}(p,q)
  +
  B
  +
  \frac12
  \Exp_{\lambda,G}^2(p,q)
  (1-2p)
  +
  Cp,
\end{multline}
where $A$, $B$, and $C$ do not depend on $(p,q)$.
Since $\dom G=\starpoundsstar$,
$
  \left|
    \Exp_{\lambda,G}(p,q)
  \right|
  \to
  \infty
$
as $p\to0$ or $p\to1$,
and so
\begin{equation}\label{eq:-infty}
  \lim_{\substack{(p,q)\to(1,0)\\(p,q)\in\starpoundsstar}}
  S_n(p,q)
  =
  -\infty
\end{equation}
and
\begin{equation}\label{eq:infty}
  \lim_{\substack{(p,q)\to(0,1)\\(p,q)\in\starpoundsstar}}
  S_n(p,q)
  =
  \infty.
\end{equation}
The \emph{stripped lexicographic ALN}
(or, briefly, $\starpoundsstar$ALN)
is defined as the lexicographic ALN
except that:
\begin{itemize}
\item
  its parameter is a forecast-continuous kernel $\KKK$
  on $\starpoundsstar\times\mathbf{X}$;
\item
  it outputs a root $(p,q)$
  (an element of $\starpoundsstar=\dom S_n$)
  of the equation $S_n(p,q)=0$
  as $(p_n,q_n)$
  and crashes if this equation does not have roots
  (this will never happen for the kernel (\ref{eq:kernel-theorem})).
\end{itemize}
Because of (\ref{eq:-infty}) and (\ref{eq:infty}),
$\starpoundsstar$ALN applied to the kernel (\ref{eq:kernel-theorem})
on $\starpoundsstar\times\mathbf{X}$
still ensures that (\ref{eq:K29}) holds
for our feature mapping $(\Phi_0,\Phi_1)$;
this algorithm never crashes and, of course, never outputs $(p_n,q_n)$
with $p_n\in\{0,1\}$.
We can see that the proof of (\ref{eq:goal})
given in the previous subsection
still works.

Let us now prove that $\CC_{\lambda,\FFF}<\infty$
when $\dom G=\starpoundsstar$.
It suffices to check that
\begin{equation}\label{eq:p}
  \limsup_{\substack{(p,q)\to(0,1)\\(p,q)\in\starpoundsstar}}
  p
  \Exp_{\lambda,G}^2(p,q)
  <
  \infty
\end{equation}
and
\begin{equation}\label{eq:1-p}
  \limsup_{\substack{(p,q)\to(1,0)\\(p,q)\in\starpoundsstar}}
  (1-p)
  \Exp_{\lambda,G}^2(p,q)
  <
  \infty.
\end{equation}
For example, let us demonstrate (\ref{eq:1-p}).
Without loss of generality,
we replace (\ref{eq:tails1}) with
\begin{equation}\label{eq:tails2}
  f'_{-}(t)=O(t^{-2}),
  \quad
  g'_{-}(t)=O(t^{-2})
\end{equation}
(this can be done since $f'_{-}(t)\le f'_{+}(t)\le f'_{-}(t+1)$
and $g'_{-}(t)\le g'_{+}(t)\le g'_{-}(t+1)$).
Consider the decision
\begin{equation*}
  (X,Y)
  :=
  \left(
    \lambda(0,G(p,q)),
    \lambda(1,G(p,q))
  \right).
\end{equation*}
Since $-\frac{1-p}{p}$ is a subgradient
(see, e.g., \citealt{rockafellar:1970}, Section 23) of $f(x)$ at $X$,
(\ref{eq:tails2}) implies that
$1-p=O(X^{-2})$,
i.e.,
$(1-p)X^2=O(1)$.
Since $\lvert\Exp_{\lambda,G}(p,q)\rvert=X-Y\le X-C_1$
for $(p,q)<(1,0)$ sufficiently close to $(1,0)$,
(\ref{eq:1-p}) indeed holds.

\subsection*{Proof: Part III}

In this subsection we consider the remaining possibilities
for $\dom G$.
Let us define the \emph{left-stripped lexicographic ALN}
($\starpounds$ALN for brief)
as the lexicographic ALN except that:
\begin{itemize}
\item
  its parameter is a forecast-continuous kernel $\KKK$
  on $\starpounds\times\mathbf{X}$,
  where the \emph{left-stripped lexicographic square}
  \begin{equation*}
    \starpounds
    :=
    (0,1]\times[0,1]
  \end{equation*}
  is equipped with the order and topology inherited from $\pounds$;
\item
  it outputs a root $(p,q)\in\starpounds$
  of the equation $S_n(p,q)=0$ as $(p_n,q_n)$;
  if this equation does not have roots in $\starpounds$,
  we set $p_n:=1$ and set $q_n\in[0,1]$ arbitrarily
  (we will make sure that this happens only when $S_n$ is everywhere positive).
\end{itemize}
In a similar way we define the \emph{right-stripped lexicographic square} $\poundsstar$
and the \emph{right-stripped lexicographic ALN}
($\poundsstar$ALN),
which always outputs $(p_n,q_n)\in\poundsstar$;
when $S_n(p,q)=0$ does not have roots $(p,q)\in\poundsstar$
we now set $p_n:=0$.

We only consider the case $\dom G=\starpounds$
(the case $\dom G=\poundsstar$ is treated analogously);
this corresponds to $f(C_0)=\infty$ and $f(C_1)<\infty$.
Since $S_n$ is continuous,
the absence of roots of $S_n=0$ in $\starpounds$
in conjunction with (\ref{eq:infty})
means that $S_n$ is positive everywhere on $\starpounds$,
and so setting $p_n:=1$ in this case
guarantees that $\starpounds$ALN still ensures (\ref{eq:K29}).
It remains to notice that (\ref{eq:p}) still holds.

\section{The \Uppercase{a}lgorithm}
\label{sec:algorithm}

In this short section we extract the decision strategy
achieving (\ref{eq:goal})
from our proof of Theorem \ref{thm:main}.
As we have already noticed (see (\ref{eq:kernel-chain})),
\begin{multline}\label{eq:S}
  S_n(p,q)
  =
  \sum_{i=1}^{n-1}
  \Bigl(
    \Exp_{\lambda,G}(p,q)
    \Exp_{\lambda,G}(p_i,q_i)
    +
    \KKK(x_n,x_i)
  \Bigr)
  (y_i-p_i)
  \\+
  \frac12
  \Bigl(
    \Exp_{\lambda,G}^2(p,q)
    +
    \KKK(x_n,x_n)
  \Bigr)
  (1-2p);
\end{multline}
this immediately leads to the following explicit description.

\bigskip

\noindent
\textsc{An algorithm achieving (\ref{eq:goal})}

\parshape=3
\IndentI   \WidthI
\IndentII  \WidthII
\IndentII  \WidthII
\noindent
\ifJOURNAL
\textbf{Parameters:}
  game with loss function $\lambda$ and canonical choice function $G$;
  kernel $\KKK$ on $\mathbf{X}$
\fi
\ifnotJOURNAL
\begin{tabbing}
\textbf{Parameters:}
  \=game with loss function $\lambda$ and canonical choice function $G$;\\
  \>kernel $\KKK$ on $\mathbf{X}$
\end{tabbing}
\fi

\parshape=9
\IndentI   \WidthI
\IndentII  \WidthII
\IndentII  \WidthII
\IndentII  \WidthII
\IndentIII \WidthIII
\IndentIII \WidthIII
\IndentII  \WidthII
\IndentII  \WidthII
\IndentI   \WidthI
\noindent
FOR $n=1,2,\dots$:\\
  Read $x_n\in\mathbf{X}$.\\
  Define $S_n(p,q)$ by (\ref{eq:S}) for all $(p,q)\in\pounds$ for which $G(p,q)$ is defined.\\
  Define $(p_n,q_n)$ as any root $(p,q)$ of $S_n(p,q)=0$;\\
    if there are no roots,\\
    set $p_n:=(1+\sign S_n)/2$ and set $q_n$ to any number in $[0,1]$.\\
  Set $\gamma_n:=G(p_n,q_n)$.\\
  Read $y_n\in\{0,1\}$.\\
END FOR.

\bigskip

\noindent
(We saw in the previous section that $\sign S_n$ is well defined
and is $-1$ or $1$ in this context.)

The canonical choice functions for the three examples of games
given in \S\ref{sec:examples} are as follows:
$G(p,q)=p$ for the square loss and log loss games,
and
\begin{equation}\label{eq:absolute-choice}
  G(p,q)
  =
  \begin{cases}
    0 & \text{if $p<1/2$}\\
    1 & \text{if $p>1/2$}\\
    q & \text{if $p=1/2$}
  \end{cases}
\end{equation}
for the absolute loss game.

\section{Directions of \Uppercase{f}urther \Uppercase{r}esearch}
\label{sec:discussion}

In this section we discuss informally
what we consider to be interesting directions of further research.

\subsection*{Non-convex \Uppercase{g}ames}

Theorem \ref{thm:main} assumes that the superdecision set is convex.
The assumption of convexity is convenient
but not indispensable.
We will only discuss the simplest non-convex game.

The loss function for the \emph{simple loss game}
is the same as for the absolute loss game,
$\lambda(y,\gamma)=\lvert y-\gamma\rvert$,
but $\Gamma=\{0,1\}$.
Now the approach we have used in this paper does not work:
since $\Gamma$ consists of two elements,
there is no non-trivial continuous choice function $G:\pounds\to\Gamma$
(every continuous image of $\pounds$ is connected:
\citealt{engelking:1989local}, Theorem 6.1.4).

A natural idea (\citealt{cesabianchi/etal:1997})
is to allow Decision Maker to use randomization.
The expected loss of a strategy making decision $1$ with probability $\gamma$
and $0$ with probability $1-\gamma$
is $\lvert y-\gamma\rvert$,
where $y$ is the actual observation;
therefore, for the simple loss game a randomized decision strategy
can guarantee the following analogue of (\ref{eq:absolute}):
\begin{equation}\label{eq:simple}
  \sum_{n=1}^N
  \Expect \lvert y_n-\gamma_n\rvert
  \le
  \sum_{n=1}^N
  \left|
    y_n-D(x_n)
  \right|
  +
  \frac{\sqrt{6}}{4}
  \left(
    \left\|2D-1\right\|_{\SSS}
    +
    1
  \right)
  \sqrt{N},
\end{equation}
where $\Expect$ refers to the strategy's internal randomization
(the decision rules $D$ can be allowed to take values in $[0,1]$).

The disadvantage of (\ref{eq:simple}) is that typically
we are interested in the strategy's actual rather than expected loss.
Our derivation of (\ref{eq:simple}) shows the role of randomization:
with our choice function (\ref{eq:absolute-choice})
no randomization is required unless $p=1/2$.
Typically, we rarely find ourselves in a situation of complete uncertainty,
$p_n=1/2$;
therefore, only a little bit of randomization is needed,
essentially for tie breaking.
The actual loss will be very close to the expected loss.
It would be interesting to derive formal statements
along these lines.

\subsection*{Non-binary \Uppercase{o}bservations}

It would also be interesting to extend this paper's results
to more general observation spaces
(first of all,
to carry them over to least-squares regression and multi-class classification).
The two apparent obstacles to such extensions
are that the fundamental equality (\ref{eq:fundamental})
looks tailored to the binary case $y\in\{0,1\}$
and that Lemma~\ref{lem:basic} ceases to be obvious outside the binary case.
However,
(\ref{eq:fundamental}) only states,
in the terminology of \citet{shafer/vovk:2001},
that $\lambda(p,\gamma)$ is the game-theoretic expected value of $\lambda(y,\gamma)$
(and that reproducing $\lambda(y,\gamma)$ given $\lambda(p,\gamma)$
can be accomplished by buying $\lambda(1,\gamma) - \lambda(0,\gamma)$ tickets
paying $y$ and costing $p$ each).
Similar equalities hold for many other forecasting protocols.
And an analogue of Lemma \ref{lem:basic}
for a wide class of forecasting protocols is proved in \citet{\GTPX}.

\subsection*{Optimality}

An important problem is to investigate
the optimality of our algorithm,
described in \S\ref{sec:algorithm}:
is the bound (\ref{eq:goal}) tight?
(The tightness of the bounds
in Theorem \ref{thm:K29} and Equation (\ref{eq:simple-resolution})
is established in \ifJOURNAL later versions of \fi\citealt{\GTPXIII}.)

\ifJOURNAL
\acks{This work was partially supported by MRC (grant S505/65)
and Royal Society.}
\fi

\ifnotJOURNAL
\subsection*{Acknowledgments}

This work was partially supported by MRC (grant S505/65)
and Royal Society.
\fi

\ifnotJOURNAL
  \bibliographystyle{plain}
\fi

\end{document}